\pdfoutput=1

\documentclass[11pt]{article}

\usepackage{acl}

\usepackage{times}
\usepackage{latexsym}
\usepackage{algorithmic}
\usepackage{tabularx}
\usepackage{array}

\usepackage{graphicx}    
\usepackage[table,xcdraw]{xcolor}
\usepackage[T1]{fontenc}

\usepackage[utf8]{inputenc}

\usepackage{microtype}

\usepackage{inconsolata}

\usepackage[ruled]{algorithm2e}
\usepackage{graphicx}
\usepackage{multirow}
\usepackage{latexsym}
\usepackage{amsmath,amssymb,amsthm}
\usepackage{subfigure}
\usepackage{microtype}
\DeclareGraphicsExtensions{.pdf, .png}

\usepackage{booktabs,arydshln}
\usepackage{setspace}
\usepackage[normalem]{ulem}
\useunder{\uline}{\ul}{}
\usepackage{tikz}
\usepackage{bbding}
\usepackage{pifont}
\usepackage{wasysym}
\usepackage{paralist}
\usepackage{setspace}
\usepackage{tikz}
\usepackage{pgf-pie} 
\usepackage{subcaption}
\usepackage{xcolor}
\usepackage{float} 
\usepackage{subcaption}
\newcommand{\result}[2]{\textbf{#1}{\tiny$\scriptstyle\pm#2$}}

\title{Beyond Invariance: Factor-wise Selective Treatment of Shifts for Multimodal Learning}

\author{Guimin Hu$^{1}$, Qiwei Peng$^{2}$, Boyan Xu$^{1}$, Ruichu Cai$^{1,3}$ \\
  $^1$Guangdong University of Technology\\
  $^2$University of Copenhagen \\
  $^3$Guangdong Provincial Key Laboratory \\of Media Convergence and Digital Creativity\\
\texttt{rice.hu.x@gmail.com}}

\begin{document}

\maketitle
\begin{abstract}
Distribution shifts pose a fundamental challenge to robust multimodal learning. Existing methods typically promote invariance by suppressing environment-dependent variations, implicitly applying a uniform treatment to shifted information. However, shifted information can have different predictive consequences: some contributes little to prediction, some remains reliably informative, and some exhibits environment-dependent predictive effects. This raises a fundamental question: which shifted information should be suppressed, preserved, or attenuated? To address this question, we propose FaShift, a factor-wise shift framework that diagnoses latent representation factors according to their distributional changes, intervention effects, and cross-environment stability. By factorizing multimodal representations and evaluating these complementary properties, FaShift distinguishes non-predictive factors, stably predictive factors, and predictively useful but unstable factors. It then selectively suppresses, preserves, or attenuates their contributions, rather than enforcing uniform invariance. Experiments on MOSI, MOSEI, and CH-SIMS show that FaShift achieves the best ID accuracy and F1 across all three datasets. Compared with the strongest competitors reporting OOD results, it improves OOD accuracy/F1 by 1.5/3.6 points on MOSI and 1.7/2.7 points on MOSEI, while achieving the best reported OOD accuracy and F1 on CH-SIMS.
\end{abstract}

\section{Introduction}
Multimodal learning exploits heterogeneous modalities to capture complementary information, but data distributions can vary across domains, users, acquisition conditions, and time. Such distribution shifts may induce reliance on environment-specific correlations, degrading generalization to unseen environments. A common approach is invariant representation learning, which aims to reduce environment-dependent variation while preserving predictive information that generalizes across environments.

\begin{figure}[t] \centering \includegraphics[width=1.01\linewidth]{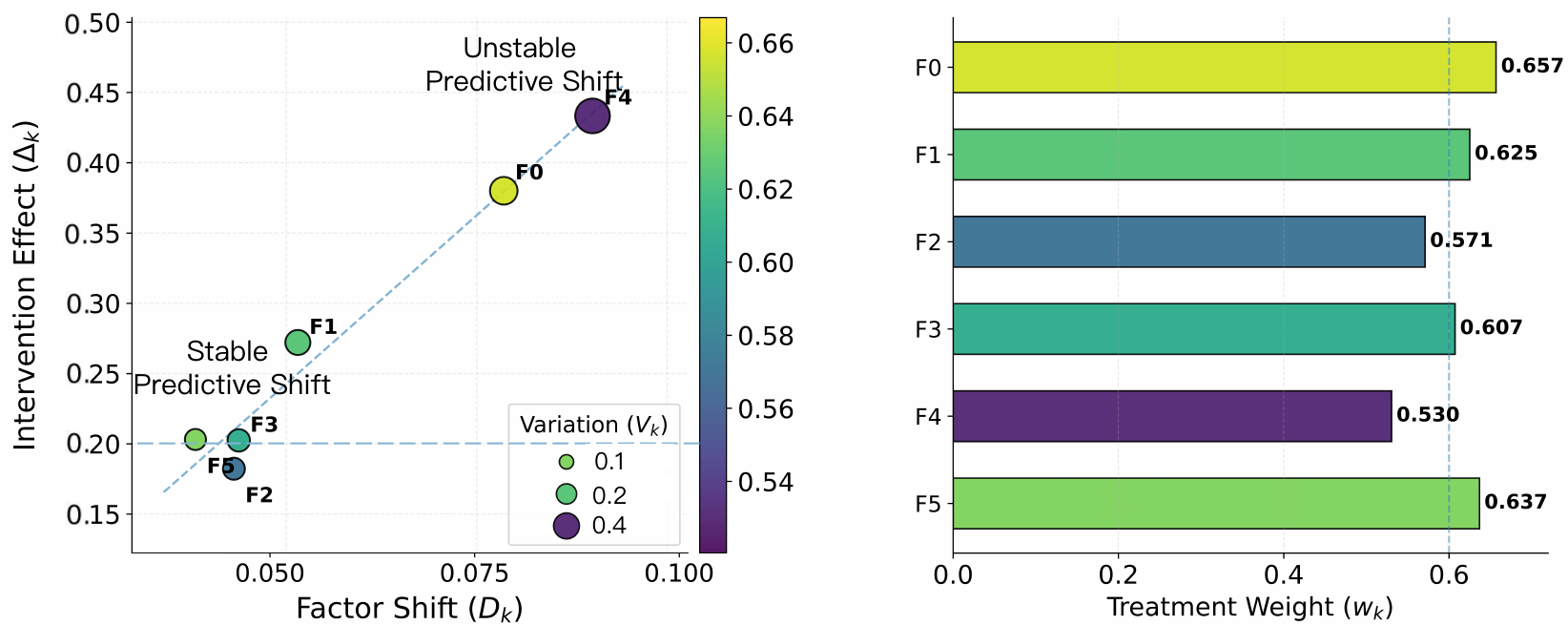} \caption{Heterogeneous factor shifts motivate shift-aware reliance. Left: Factor-wise shift and intervention effects. Right: Shift-aware factor reliance.} \label{fig:teaser} \end{figure}

However, environment-dependent variation is not necessarily spurious. Shifted representations may retain predictive information, yet exhibit different predictive consequences across factors. Thus, the key challenge extends beyond identifying invariant information to assessing the predictive effects of shifted information and their stability across environments. This raises a fundamental question: \emph{how should shifted information be treated according to its predictive effects and cross-environment stability?}



To investigate this question at the factor level, we decompose multimodal representations into latent factors. As illustrated in Figure~\ref{fig:teaser}, these factors vary substantially in shift magnitude, predictive effects, and the cross-environment stability of those effects (left). Notably, similar shifts can induce markedly different changes in model predictions (right), indicating that shift magnitude alone is insufficient to determine how a factor should be treated. This motivates \emph{factor-wise diagnosis}, which determines whether each shifted factor should be suppressed, preserved, or attenuated based on its predictive effects and their cross-environment stability.

Existing approaches address distribution shifts through invariant representation learning, causal adaptation, or evolving domain generalization~\cite{mai2026learning,zhong2025towards,he2025learning}. Although these methods account for the potential predictive value of environment-dependent information, they generally emphasize learning transferable representations or adaptation mechanisms, with less focus on systematically diagnosing how individual shifted factors affect predictions and whether these effects remain stable across environments~\cite{sun2022counterfactual,mai2023cmib}.

To address this question, we propose \emph{FaShift} (\emph{Factor-wise Shift Discovery}), a factor-level framework for diagnosing and selectively adapting representation shifts. FaShift first decomposes multimodal representations into latent factors and characterize each factor from three complementary perspectives: \emph{distributional shift}, \emph{interventional effect}, and \emph{cross-environment stability}. Distributional shift measures how a factor's representation changes across environments. We then intervene on each factor while keeping the remaining factors fixed and quantify the resulting change in prediction. Finally, we examine whether this interventional effect remains consistent across environments. These three characteristics distinguish \emph{whether a factor shifts}, \emph{whether it affects prediction}, and \emph{whether its predictive effect is stable across environments.}

Based on these factor-wise characteristics, FaShift performs \emph{selective adaptation} rather than enforcing universal invariance. Factors with negligible interventional effects are suppressed, while factors with meaningful and stable effects are retained. For factors with meaningful but unstable effects, FaShift attenuates their contributions according to their degree of cross-environment instability. In this way, FaShift identifies which shifted factors are consequential for prediction and determines whether their contributions should be \emph{suppressed, preserved, or attenuated} according to the stability of their effects across environments. This factor-wise treatment provides a shift-aware alternative to universal invariance for robust multimodal learning. Our contributions are summarized as follows:
\begin{itemize}
\item We show that shift magnitude alone is insufficient to determine factor treatment: factors with similar distributional shifts can have markedly different predictive consequences, motivating selective rather than uniform treatment of shifted information.

\item We propose \textbf{FaShift}, a factor-wise framework that diagnoses individual representation factors through distributional shift, intervention effect, and cross-environment stability, and accordingly suppresses, preserves, or attenuates their contributions.

\item Extensive experiments across multimodal datasets show that FaShift consistently improves \textbf{OOD generalization} while maintaining \textbf{SOTA ID performance}, demonstrating the effectiveness of selective treatment over universal invariance.
\end{itemize}

\section{Related Work}

\subsection{Invariant Multimodal Learning}

Distribution shifts have motivated invariant representation learning and domain generalization, which seek predictive relationships that remain stable across environments. Recent multimodal studies have incorporated invariant perspectives to improve robustness under such shifts. Domain Generalization via Causal Adjustment~\cite{wang2024domain} employs backdoor adjustment to remove domain-specific confounding and learn domain-invariant representations. CmIR~\cite{mai2026learning} disentangles modality representations into causal-invariant and environment-specific components, while SCLF~\cite{cao2026sparse} identifies stable causal latent features through invariance and stability constraints. Other approaches explore counterfactual intervention for robust emotion recognition~\cite{zhong2025towards} and causal representation learning for domain generalization~\cite{liang2026multimodal}. These methods primarily seek invariant or stable representations by suppressing environment-dependent variations. In contrast, we ask how each shifted factor should be treated according to its intervention effect and cross-environment stability.

\subsection{Distribution Shifts and Adaptation}

Beyond invariant representation learning, causal adaptation studies have examined how predictive mechanisms transfer or evolve across environments. Causal domain adaptation methods model domain differences as changes in the underlying data-generating process and seek relationships that remain transferable across domains~\cite{magliacane2018domain,teshima2020few,yue2021transporting}. More recent work on evolving domain generalization further recognizes that environment-dependent factors may contain useful information, motivating the joint modeling of invariant and dynamic representations~\cite{xie2024enhancing,he2025learning}. These studies highlight that distribution shifts need not uniformly correspond to spurious variation. However, existing approaches primarily focus on transferring mechanisms or learning representations that generalize across evolving environments, without explicitly characterizing how individual factors change or whether such changes affect prediction. We instead take a shift-centric perspective, quantifying factor-wise shifts and intervention effects to determine which should be suppressed or preserved, rather than uniformly enforcing invariance.

\section{FaShift}

\subsection{Problem Formulation}

Given a multimodal observation $\mathcal{X}$ with label $Y$ under an environment $E\in\mathcal{E}$, we aim to determine how environment-induced representation shifts should be treated for robust prediction. While conventional robust learning often seeks to suppress environment-dependent variations, such a strategy implicitly assumes that all shifts are undesirable. This raises a fundamental question: \textit{when a representation shifts across environments, should the shift be suppressed, preserved, or adaptively attenuated?}

Each environment represents a distinct data-generating condition, such as domain, population, acquisition setting, or temporal context. The key challenge is therefore to distinguish representation shifts that contribute to prediction from those that do not, and to determine their appropriate treatment across environments.

\paragraph{Shift Does Not Directly Determine Treatment.}
An environment-induced shift does not by itself determine how the corresponding factor should be treated. A factor may vary substantially across environments while having little effect on the prediction when intervened upon. Therefore,
\begin{equation}
D_k^{a,b}>0
;\not\Rightarrow;
|\Delta_k^{a\rightarrow b}|>0.
\end{equation}

Conversely, a shifted factor may exhibit a substantial intervention-based effect on the prediction:
\begin{equation}
D_k^{a,b}>0
\quad\text{and}\quad
|\Delta_k^{a\rightarrow b}|>0.
\end{equation}

Thus, shift magnitude alone cannot determine factor treatment. Intervention effects indicate prediction relevance, while cross-environment stability determines whether relevant factors should be preserved or attenuated. This motivates our factor-wise analysis for selective treatment.

\begin{figure}[t]
    \centering
    \includegraphics[width=\linewidth]{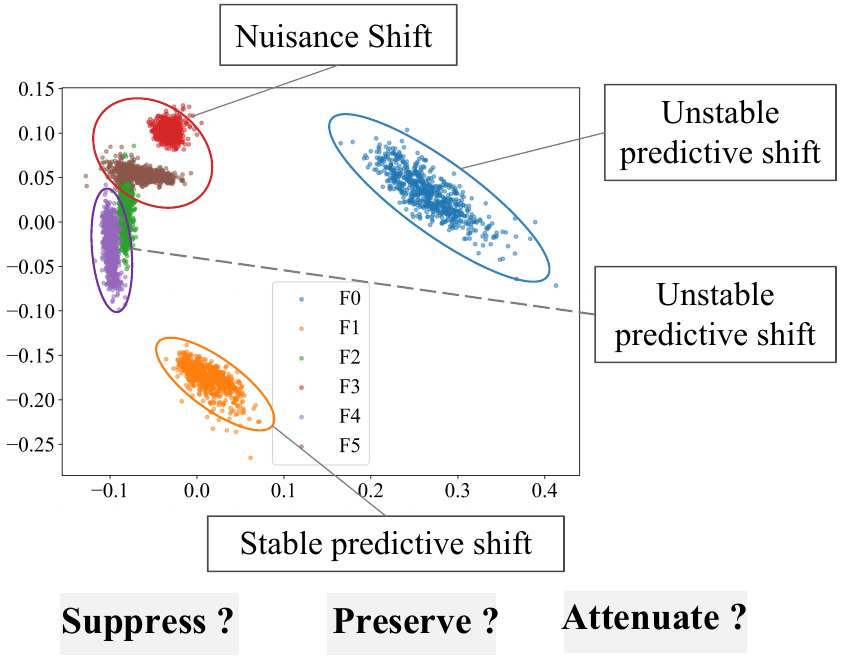}
    \caption{Visualization of factor-wise shift diagnosis in the PCA space of the learned latent factors. FaShift identifies heterogeneous shift patterns across latent factors, distinguishing nuisance, unstable predictive, and stable predictive shifts.}
    \label{fig:method_teaser}
\end{figure}

\subsection{Stage I: Factor Discovery.}

\paragraph{Factorized Representation Learning.}
We first encode and fuse the multimodal inputs into a joint representation using a multimodal encoder $f_{\mathrm{mm}}$: $\mathbf{H}=f_{\mathrm{mm}}(\mathcal{X})$, which is subsequently factorized into $K$ latent factors:
\begin{equation}
\mathbf{Z}
=
[\mathbf{z}_1,\ldots,\mathbf{z}_K]
=
\mathcal{F}_{\theta}(\mathbf{H}).
\end{equation}
where $\mathcal{F}_{\theta}$ denotes the factorization module and each $\mathbf{z}_k$ captures a potentially distinct aspect of the predictive representation without requiring predefined semantic annotations. The resulting factorized representation is then mapped to the task output by a prediction head $g(\cdot)$:

\begin{equation}
\hat{Y}_{Z}=g(\mathbf{Z}).
\end{equation}
To reduce redundancy among factors, we minimize their pairwise
statistical dependence using the Hilbert-Schmidt Independence
Criterion (HSIC)\cite{gretton2005measuring}:
\begin{equation}
\mathcal{L}_{\mathrm{dec}}
=\sum_{i\neq j}\operatorname{HSIC}(\mathbf{z}_i,\mathbf{z}_j),
\end{equation}

where HSIC measures statistical dependence between two random
variables. The representation objective is
\begin{equation}
\mathcal{L}_{\mathrm{rep}}
=\ell(\hat{Y}_{Z},Y)
+
\lambda_{\mathrm{dec}}\mathcal{L}_{\mathrm{dec}}.
\end{equation}
where \(\ell(\cdot,\cdot)\) denotes the task-specific prediction loss. This stage learns a non-redundant factorized representation that preserves task-relevant information and supports factor-level intervention, without assuming environmental stability of individual factors. We assess factor redundancy by computing the Pearson correlation between the $\ell_2$-norm activations of each factor pair. Lower correlations indicate less overlap between factors (see Experiment section). The factorization is designed to preserve the task-relevant information in the original representation, such that the factorized representation $\mathbf{Z}$ retains predictive capability for $Y$. Specifically, we require a small gap between $\hat{Y}_Z$ and $\hat{Y}_H=g(\mathbf{H})$ using the same prediction head.

\subsection{Stage II: Factor-wise Shift Diagnosis}

We characterize each factor by its distributional shift, intervention-based predictive relevance, and cross-environment stability. Given a source environment $E_a$ and a target environment $E_b$, factor-level interventions assess the predictive effects of cross-environment changes, while cross-modal factor interactions capture their joint effects. These statistics guide factor-wise treatment under distributional shifts.

\paragraph{Environment Construction.}
We construct training environments to capture heterogeneous conditions for factor-wise shift diagnosis. When reliable environmental attributes are available, such as speaker, participant, session, or temporal context, we use them directly to define environments. Otherwise, we construct pseudo-environments from the Stage-I factorized representations by computing the factor activation profile.
\begin{equation}
\mathbf{s}^{(n)}
=[
|\mathbf{z}_1^{(n)}|_2,
\ldots,
|\mathbf{z}_K^{(n)}|_2
]
\end{equation}
for each training sample and cluster these profiles into groups using $K$-means. The resulting groups serve as pseudo-environments for shift diagnosis. Environment pairs $(E_a,E_b)$ are then used to assess factor-wise distributional shifts, intervention effects, and cross-environment stability. Environments are constructed using training data only, with validation data used for hyperparameter selection and the test set reserved for final evaluation.

\paragraph{Shift Discovery.}
For each factor $\mathbf{z}_k$, we measure its distributional shift between an environment pair $(E_a,E_b)$ by
\begin{equation}
D_k=
\mathcal{D}\big(
P(\mathbf{z}_k|E_a),
P(\mathbf{z}_k|E_b)
\big),
\end{equation}
where $\mathcal{D}$ denotes a distribution discrepancy measure such as Maximum Mean Discrepancy (MMD) or Wasserstein distance. A factor is regarded as shifted when $D_k>\tau_s$.

\begin{table}[t]
\centering
\small
\setlength{\tabcolsep}{3pt}
\renewcommand{\arraystretch}{1.08}
\begin{tabular}{lcc}
\toprule
\textbf{Shift Type} & \textbf{Decision Criterion} & \textbf{Treatment} \\
\midrule
Nuisance Shift
& $\Delta_k \leq \tau_c$
& \textbf{Suppress} \\

Stable Predictive Shift
& $\Delta_k > \tau_c,\; V_k \leq \tau_v$
& \textbf{Preserve} \\

Unstable Predictive Shift
& $\Delta_k > \tau_c,\; V_k > \tau_v$
& \textbf{Attenuate} \\
\bottomrule
\end{tabular}
\caption{Shift categorization and corresponding selective treatment of representation factors. $\tau_c$ and $\tau_v$ are the thresholds for intervention effect and instability, respectively.}
\label{tab:causal_factor}
\end{table}

\paragraph{Factor-level Intervention Attribution.}
For a sample from environment \(E_b\), we treat \(\mathbf{z}_k\) as the factor to be intervened on and identify the top-\(M\) nearest counterparts in a reference environment \(E_a\) by matching the non-intervened factors \(\mathbf{z}_{-k}\). We then average their corresponding \(\mathbf{z}_k\) representations to obtain a reference factor \(\mathbf{z}_k^{a,*}\), which is used to construct the counterfactual representation:
\begin{equation}
\mathbf{z}^{k\leftarrow a}=
[\mathbf{z}_1^b,\ldots,\mathbf{z}_k^{a,*},\ldots,\mathbf{z}_K^b].
\end{equation}
Here, only $\mathbf{z}_k$ is replaced, while all non-intervened factors $\mathbf{z}_{-k}$ remain unchanged. We measure the similarity between samples based on their non-intervened factors using cosine similarity, and select the top-\(M\) nearest counterparts from the reference environment.
The relevance of the shift is quantified by the prediction change:
\begin{equation}
\Delta_k^{a\rightarrow b}
=
d_Y\bigl(g(\mathbf{z}^b),g(\mathbf{z}^{k\leftarrow a})\bigr),
\end{equation}
where $d_Y$ measures the difference between the factual and
counterfactual predictions.

\paragraph{Joint Intervention Attribution.}
Factor-level intervention alone may not fully capture the joint effects of multimodal factors, whose combined influence can be complementary, redundant, or conflicting. We therefore further consider pairwise joint interventions to characterize cross-factor dependencies. Specifically, for a factor pair $(\mathbf{z}_i,\mathbf{z}_j)$, we simultaneously replace both factors with their reference counterparts:
\begin{equation}
\mathbf{z}^{ij \leftarrow a}
=[
\mathbf{z}_1^b,\ldots,
\mathbf{z}_i^{a,*},\ldots,
\mathbf{z}_j^{a,*},\ldots,
\mathbf{z}_K^b
]
\end{equation}
The joint intervention effect is defined as
\begin{equation}
\Delta_{ij}^{a \rightarrow b}
=
d_Y\bigl(
g(\mathbf{z}^b),
g(\mathbf{z}^{ij \leftarrow a})
\bigr)
\end{equation}
We then quantify the cross-factor interaction as the deviation of the joint effect from the sum of the individual effects:
\begin{equation}
I_{ij}^{a \rightarrow b}
=
\Delta_{ij}^{a \rightarrow b}
-
\Delta_i^{a \rightarrow b}
-
\Delta_j^{a \rightarrow b}.
\end{equation}
A positive $I_{ij}$ indicates a synergistic interaction, while a negative value indicates a sub-additive interaction caused by redundancy, saturation, or conflicting effects. We analyze only cross-modal factor pairs to characterize their complementary, redundant, or conflicting dependencies.

\paragraph{Cross-environment Stability.}
We further examine whether the intervention effects remain consistent
across environment pairs. For factor-level effects, we quantify
relative variability as
\begin{equation}
V_k
=
\frac{
\operatorname{Std}(\Delta_k)
}{
\operatorname{Mean}(\Delta_k)+\epsilon
}.
\end{equation}
Similarly, we measure the cross-environment instability of factor interactions as
\begin{equation}
V_{ij}
=
\frac{
\operatorname{Std}(I_{ij})
}{
\operatorname{Mean}(|I_{ij}|)+\epsilon
},
I_k = \sum_{j\neq k} V_{kj},
\end{equation}
A smaller $V_k$ or $I_{k}$ indicates greater cross-environment
stability. Based on $(D_k,\Delta_k,V_k)$ and the associated
cross-factor interaction profiles, shifted factors are categorized
and treated as summarized in Table~\ref{tab:causal_factor} and Figure~\ref{fig:method_teaser}.

\subsection{Stage III: Selective Shift Treatment}
\label{sec:selective_causal_treatment}

Based on the statistics obtained in Stage II, we determine the retention weight of each factor according to its intervention effect and cross-environment stability. Specifically, we define
\begin{equation}
w_k =
\begin{cases}
0, & \Delta_k \leq \tau_c,\\
1, & \Delta_k > \tau_c,\ V_k \leq \tau_v,\\
\displaystyle
\frac{\bar{\Delta}_k}
{\bar{\Delta}_k+\lambda_v S_k+\epsilon},
& \Delta_k > \tau_c,\ V_k > \tau_v,
\end{cases}
\label{eq:treatment_weight}
\end{equation}
where $S_k=\bar{V}_k+\lambda_{I}\bar{I}_k$ aggregates the instability and interaction terms for factor $k$, $w_k\in[0,1]$ is the retention weight of factor $k$, $\tau_c$ and $\tau_v$ are the thresholds for intervention effect and instability, respectively, $\bar{\Delta}_k$,  $\bar{I}_k$ and $\bar{V}_k$ are the normalized intervention effect and instability. $\lambda_v$ controls the penalty on instability.

The resulting retention weights are applied to the factorized representation as
\begin{equation}
\widetilde{Z}_k = w_k Z_k,
\qquad
\widetilde{Z}
=[\widetilde{Z}_1,\ldots,\widetilde{Z}_K],
\end{equation}
with the final prediction given by $\widehat{Y}=g(\widetilde{Z})$.

This selective treatment suppresses irrelevant shifts, preserves stable and task-relevant factors, and attenuates environmentally sensitive ones. The weights are derived directly from intervention-based statistics across training environments, without factor-level semantic annotations or an additional domain adaptation network.

\begin{table*}[t]
\centering
\small
\setlength{\tabcolsep}{1.5pt}
\begin{tabular}{l|cc|cc|cc|cc|cc|cc}
\toprule
& \multicolumn{4}{c|}{\textbf{MOSI}}
& \multicolumn{4}{c|}{\textbf{MOSEI}}
& \multicolumn{4}{c}{\textbf{CH-SIMS}} \\
\cmidrule(lr){2-5}
\cmidrule(lr){6-9}
\cmidrule(lr){10-13}
\textbf{Method}
& \multicolumn{2}{c|}{ID}
& \multicolumn{2}{c|}{OOD}
& \multicolumn{2}{c|}{ID}
& \multicolumn{2}{c|}{OOD}
& \multicolumn{2}{c|}{ID}
& \multicolumn{2}{c}{OOD} \\
& Acc.$\uparrow$ & F1$\uparrow$
& Acc.$\uparrow$ & F1$\uparrow$
& Acc.$\uparrow$ & F1$\uparrow$
& Acc.$\uparrow$ & F1$\uparrow$
& Acc.$\uparrow$ & F1$\uparrow$
& Acc.$\uparrow$ & F1$\uparrow$ \\
\midrule

Self-MM
& 84.9 & 84.8 & 76.7 & 76.7
& 85.2 & 85.2 &74.7 & 74.3
& 80.7 & 80.8 & 70.1* & 70.1* \\

C-MIB
& 87.8 & 87.8 & 77.9 & 77.8
& 86.9 & 86.8 & 75.1 & 75.3
& - & - & - & - \\

DMD
& 84.3 & 84.3 & 77.5 & 77.4
& 84.6 & 84.6 & 75.2 & 75.7
& - & - & - & - \\

DEVA
& 86.3 & 86.3 & - & -
& 86.1 & 86.2 & - & -
& 79.6 & 80.3 & 69.7* & 69.8* \\

AtCAF
& 88.6 & 88.5 & 78.1 & 77.2
& 87.0 & 86.8 & 75.3 & 75.7
& - & - & - & - \\

EMOE
& 84.8 & 84.8 & 77.6 & 76.8
& 85.0 & 85.0 & 75.0 & 74.6
& 78.8* & 78.5* & 67.3* & 67.3* \\

Boosting
& 88.5 & 88.4 & - & -
& 86.5 & 86.5 & - & -
& - & - & - & - \\

ITHP
& 88.5 & 88.5 & - & -
& 87.1 & 87.1 & - & -
& - & - & - & - \\

DLF
& 88.7 & 88.6 & 79.7 &77.5
& 87.5 & 87.4 & 76.8 & 76.2
& 80.4* & 80.3* & 71.2* & 70.8* \\

CaReFlow
& 89.8 & 89.7 & - & -
& 87.9 & 88.0 & - & -
& - & - & - & - \\

CmIR
& 89.6 & 89.5 & - & -
& 87.8 & 87.7 & - & -
& - & - & - & - \\

\textbf{FaShift(Ours)}
& \result{90.9}{0.3} & \result{91.4}{0.3}
& \result{81.2}{0.1} & \result{81.4}{0.2}
& \result{88.2}{0.3} & \result{88.3}{0.3}
& \result{78.5}{0.1} & \result{78.4}{0.1}
& \result{83.3}{0.2} & \result{83.3}{0.3}
& \result{73.7}{0.4} & \result{74.2}{0.2} \\

\bottomrule
\end{tabular}
\caption{Comparison of different methods under in-domain (ID) and out-of-distribution (OOD) settings on MOSI, MOSEI, and CH-SIMS. We report binary accuracy (Acc.) and F1 score. Results marked with an asterisk (*) are reproduced using the authors' publicly available code, while ``--'' indicates that results and codes are unavailable. Best results are highlighted in bold.}
\label{tab:main_results}
\end{table*}

\section{Experiments}
\subsection{Environment Setting}
We construct training environments from the natural data-generating structure of each dataset to expose the model to heterogeneous conditions for factor-wise shift diagnosis. We consider two representative types: \textbf{Human-style environments}, where each speaker or participant defines an environment and captures individual-specific multimodal characteristics; and \textbf{Temporal-style environments}, where samples are partitioned by recording time or session to capture temporal variations in multimodal characteristics. See Appendix~\ref{app} for further details on environment construction.

Unlike the environments defined for factor-level shift diagnosis and treatment policy learning, OOD settings refer to distribution-shift scenarios used exclusively for evaluation. This separation allows us to assess whether the treatment policy learned from naturally occurring training heterogeneity can generalize beyond the environments used for diagnosis.

\paragraph{Datasets.}
We evaluate our method under two settings: \textbf{in-distribution (ID)} and \textbf{out-of-distribution (OOD)}. For ID evaluation, we follow the standard train-dev-test splits of \textbf{MOSEI} \cite{zadeh2018mosei}, \textbf{MOSI} \cite{zadeh2016mosi}, \textbf{CH-SIMS} \cite{yu2020ch} and \textbf{MUD3} \cite{wang2025beyond}, where the training and test sets are drawn from the same underlying distribution. For OOD evaluation, we adopt the distribution-shifted settings introduced by Sun et al.~\cite{sun2022counterfactual} and Zhang et al.~\cite{zhang2025multimodal} on the corresponding datasets. In our experiments, we define environments based on speakers for MOSI, MOSEI, and CH-SIMS, with each speaker constituting one environment. For MU3D, we define temporal environments according to the recording period and more results are reported in Appendix \ref{app:more_results}.

\paragraph{Metric.} We use Acc-2 (Acc.) and F1 as the primary metrics, measuring classification accuracy and the balance between precision and recall, respectively.

\paragraph{Multimodal Baselines.}
We compare FaShift with representative multimodal learning methods, including Self-MM~\cite{yu2021selfmm}, C-MIB~\cite{mai2023cmib}, DMD~\cite{li2023dmd}, DEVA~\cite{wu2025deva}, AtCAF~\cite{huang2025atcaf}, EMOE~\cite{fang2025emoe}, Boosting~\cite{mai2024multimodalboosting}, ITHP~\cite{xiao2024ithp}, DLF~\cite{wang2025dlf}, CaReFlow~\cite{mai2026careflow}, and CmIR~\cite{mai2026learning}, covering diverse multimodal representation, interaction, fusion, and optimization strategies.
\begin{table}[t]
\centering
\small
\setlength{\tabcolsep}{3.5pt}
\begin{tabular}{lcccc}
\toprule
\textbf{Rep.}
& \multicolumn{2}{c}{\textbf{MOSEI}}
& \multicolumn{2}{c}{\textbf{CH-SIMS}} \\
\cmidrule(lr){2-3}
\cmidrule(lr){4-5}
& \textbf{Acc.} & \textbf{F1}
& \textbf{Acc.} & \textbf{F1} \\
\midrule
Original $\mathbf{H}$
& 84.4 & 84.2 & 75.6 & 75.7 \\
Factorized $\mathbf{Z}$
& 84.6 & 84.3 & 75.9 & 75.9 \\
\midrule
Gap
& 0.2 & 0.1 & 0.3 & 0.2 \\
\bottomrule
\end{tabular}
\caption{Task information preservation in the factorized representation.}
\label{tab:task_information_preservation}
\end{table}

\subsection{Implementation Details}
All experiments are implemented in PyTorch and conducted on a single NVIDIA GeForce RTX 4080 GPU. Unless otherwise specified, the factorized representation contains $K=6$ factors. The factorization module is optimized using AdamW with a learning rate of $4\times10^{-4}$, a weight decay of $2\times10^{-5}$, and a batch size of 128. The coefficient of the decorrelation loss is set to $\lambda_{\mathrm{dec}}=500$. For intervention-based shift analysis, we use the factor representations learned from the training data and construct reference samples from the corresponding environments. We set \(M=8\), \(\tau_c=0.2\), \(\tau_s=0.02\) and \(\tau_v=0.6\), with all hyperparameters selected on the validation set. The prediction head and the factorization module are trained using the training split only, while all OOD data are reserved for evaluation. See Appendix \ref{app:more_impl} for more implementation details.

\begin{figure}[t]
    \centering
    \includegraphics[width=\linewidth]{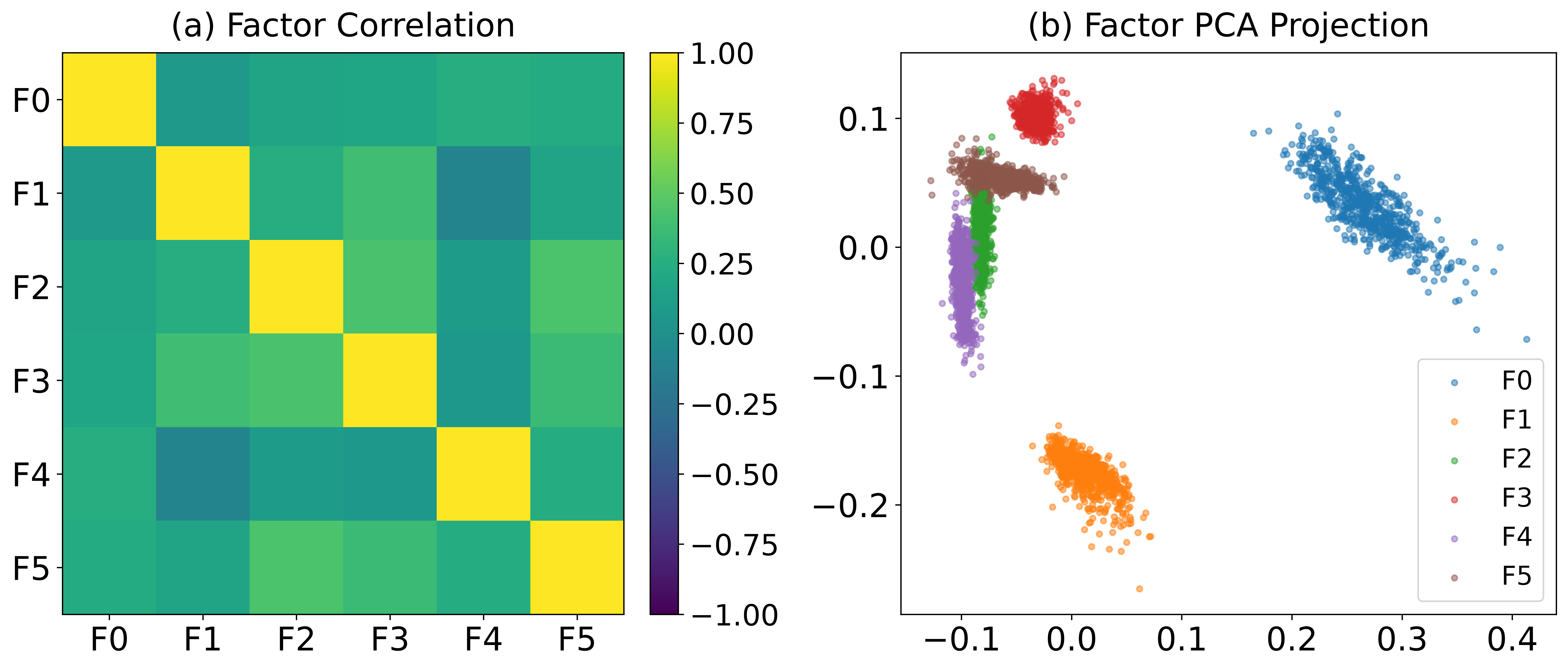}
    \caption{Visualization of the learned factors. Left: pairwise correlation matrix. Right: PCA projection of the learned factors. }
    \label{fig:stageI}
\end{figure}

\subsection{ID and OOD Results}
Table~\ref{tab:main_results} compares FaShift with existing multimodal sentiment analysis methods under both in-distribution (ID) and out-of-distribution (OOD) settings. For MOSI, MOSEI, and CH-SIMS, we use speaker-based environments to characterize factor-wise shifts across speakers. FaShift achieves the best overall performance across all three datasets, with more pronounced gains under OOD evaluation.
These results suggest that modeling factor-wise shifts provides benefits beyond conventional ID fitting. FaShift characterizes shifts across speaker-defined environments and uses intervention effects and cross-environment stability to selectively suppress, preserve, or attenuate individual factors. This enables the model to retain predictive variations while reducing the influence of unstable or less relevant shifts, which contributes to stronger OOD robustness.

The consistent gains across the three datasets support FaShift’s premise that robust multimodal learning should distinguish heterogeneous factor-wise shifts and treat them selectively, achieving a better balance between predictive performance and robustness.

\begin{table}[t]
\centering
\small
\setlength{\tabcolsep}{4pt}
\begin{tabular}{llcc}
\toprule
\textbf{Setting} & \textbf{Variant} & \textbf{Acc.} & \textbf{F1} \\
\midrule
\multirow{4}{*}{ID: MOSEI}
& w/o Factorization & 84.4 & 84.2 \\
& w/o Shift Diagnosis & 84.6 & 84.3\\
& Uniform Treatment & 83.8 & 83.6\\
& \textbf{FaShift} & \textbf{88.2} & \textbf{88.3}\\
\midrule
\multirow{4}{*}{ID: CH-SIMS}
& w/o Factorization & 75.6 & 75.7 \\
& w/o Shift Diagnosis & 75.9 & 75.9 \\
& Uniform Treatment & 75.5 & 75.1\\
& \textbf{FaShift} & \textbf{83.3} & \textbf{83.3} \\
\midrule
\multirow{4}{*}{OOD: CH-SIMS}
& w/o Factorization & 69.4 & 69.1 \\
& w/o Shift Diagnosis & 70.3& 70.5\\
& Uniform Treatment & 70.5 & 70.8\\
& \textbf{FaShift} & \textbf{73.7} & \textbf{74.2} \\
\midrule
\multirow{4}{*}{OOD: MOSEI}
& w/o Factorization & 73.9& 73.3 \\
& w/o Shift Diagnosis & 74.1 &74.2 \\
& Uniform Treatment & 74.1 & 74.6 \\
& \textbf{FaShift} & \textbf{78.5} & \textbf{78.4} \\
\bottomrule
\end{tabular}
\caption{Ablation study of FaShift under in-domain (ID) and distribution-shifted (OOD) settings.}
\label{tab:ablation}
\end{table}

\subsection{Analysis of Learned Factors}

\begin{figure}[t]
    \centering
    \includegraphics[width=\linewidth]{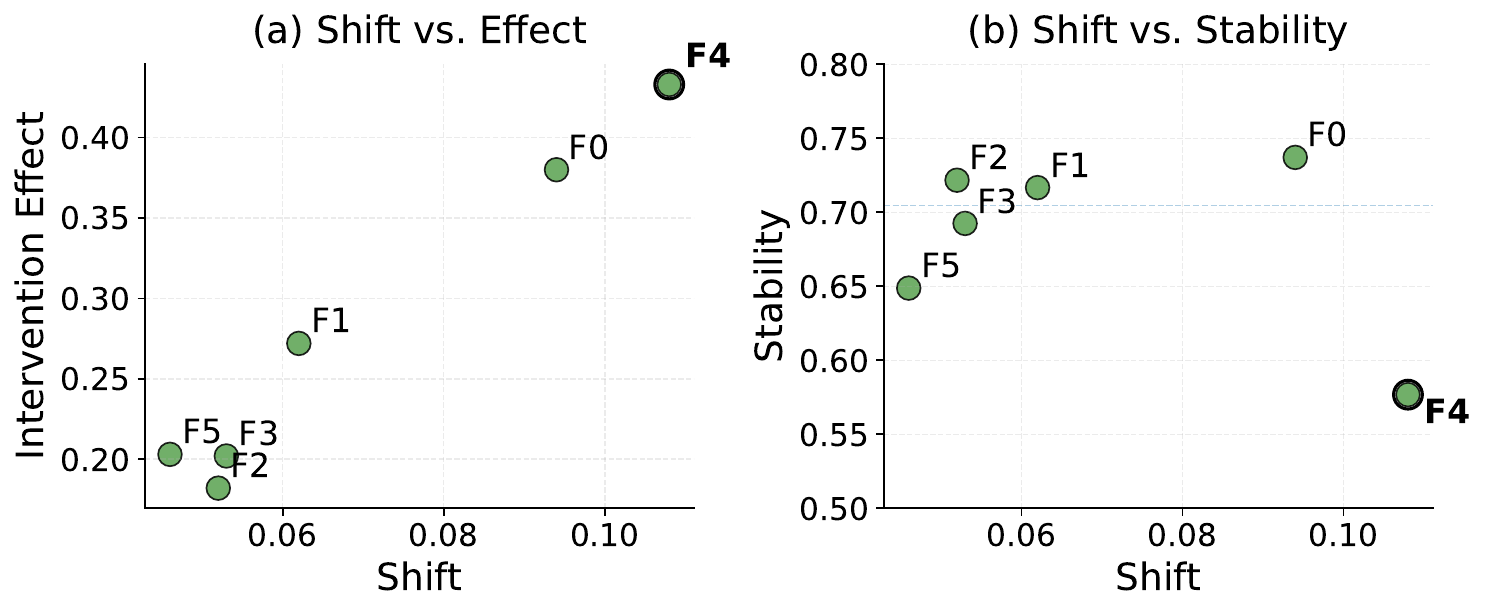}
    \caption{Factor-wise intervention effects and stability as a function of distribution shifts.}
    \label{fig:factor_visualization}
\end{figure}

\begin{figure}[t]
    \centering
    \includegraphics[width=\linewidth]{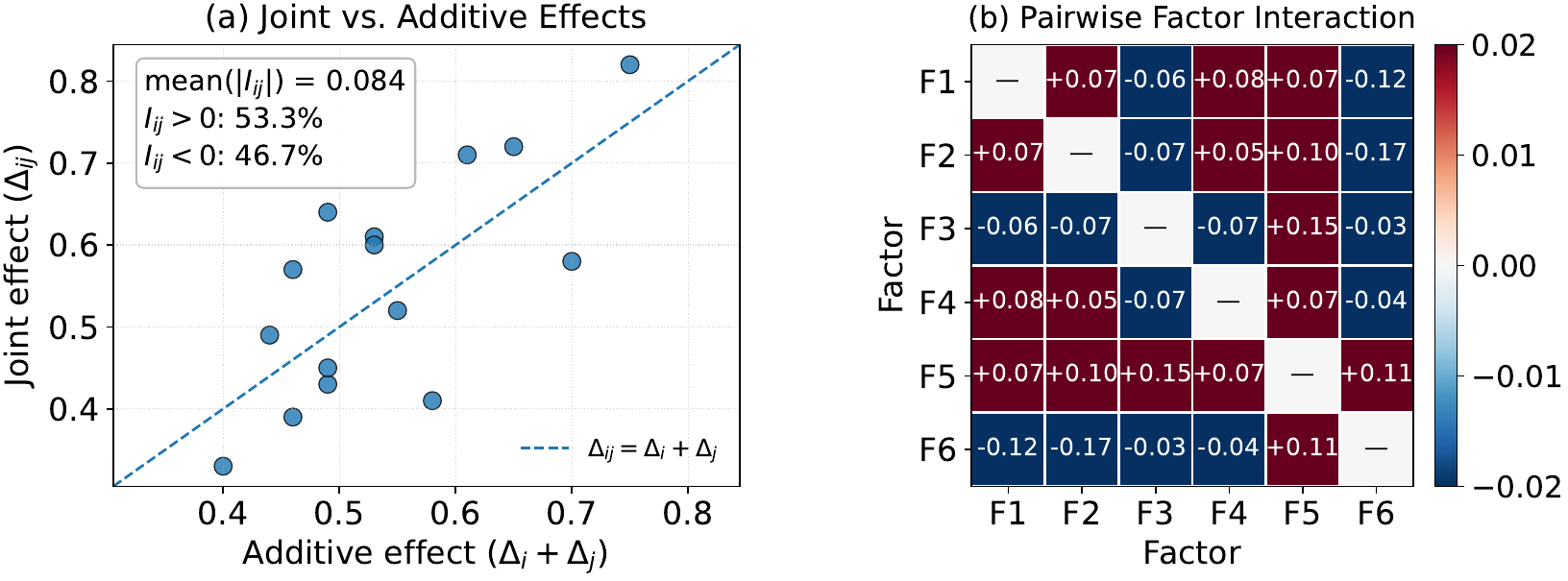}
    \caption{Pairwise joint intervention effects under distribution shifts. Left: Joint effects versus additive expectations. Right: Pairwise interaction effects, with positive and negative values indicating reinforcement and antagonism, respectively. }
    \label{fig:joint_factor}
\end{figure}

We examine whether the proposed factorization preserves task-relevant information. As shown in Table~\ref{tab:task_information_preservation}, the factorized representation $\mathbf{Z}$ achieves performance close to the original representation $\mathbf{H}$ on both CH-SIMS and MOSEI, with gaps of only 0.2 and 0.1 points in Acc. and F1 on CH-SIMS, and 0.3 and 0.2 points on MOSEI, respectively. These small differences indicate that factorization incurs minimal loss of predictive information, while $\mathbf{Z}$ retains the task-relevant information in $\mathbf{H}$ for subsequent factor-level shift analysis.

We further visualize the learned factors in Figure~\ref{fig:stageI}. The PCA projection shows that the factors exhibit structured distributions in the latent space, while the pairwise correlation matrix indicates relatively weak correlations among most factor pairs. Together, these observations suggest that the learned factors capture complementary information with limited redundancy. Importantly, these analyses serve as empirical validation of the learned representation rather than direct evidence of factor independence. Combined with the task information preservation results, they support the use of $\mathbf{Z}$ as the basis for the subsequent factor-level shift analysis.

\subsection{Ablation Study}
Table~\ref{tab:ablation} presents an ablation study of FaShift under both ID and OOD settings. Removing factorization causes a substantial performance drop across all datasets. In this variant, the entire multimodal representation is treated as a single factor. Removing shift diagnosis further degrades performance, indicating that factorization alone is insufficient without characterizing factor-wise shift relevance and cross-environment stability. This variant directly preserves all learned factors, i.e., $w_k=1$ for all factors, without intervention-based shift diagnosis. Finally, Uniform Treatment retains both factorization and shift diagnosis but applies a shared treatment weight, i.e., $w_k=\bar{w}$ for all factors. Its inferior OOD performance highlights the importance of factor-wise adaptive treatment under heterogeneous shifts.

Uniform Treatment outperforms the two ablated variants but remains consistently inferior to FaShift, particularly under OOD settings. This demonstrates that different factors require different treatments based on their intervention effects and stability. Overall, the results validate the complete factor-wise shift treatment pipeline, with factorization enabling fine-grained analysis, shift diagnosis identifying heterogeneous shifts, and adaptive treatment selectively retaining factors according to their properties.

\subsection{Factor Shift Analysis}

As shown in Figure~\ref{fig:factor_visualization}, factors exhibit substantially different intervention effects and cross-environment stability under comparable levels of shift. For example, $F_4$ experiences the largest shift and also shows the strongest intervention effect, but its effect is less stable across environments. In contrast, $F_0$ exhibits a relatively large shift and a strong intervention effect while maintaining higher stability. Other factors, such as $F_2$ and $F_3$, undergo smaller shifts but show close intervention effects and high cross-environment stability. Based on their intervention effects and cross-environment stability, our framework classifies ${F_1,F_3,F_5}$ as stable predictive factors, $F_2$ as a nuisance factor, and ${F_0,F_4}$ as unstable predictive factors, thereby guiding their selective treatment.

We further investigate whether the effects of individual factors combine additively. Figure~\ref{fig:joint_factor} compares pairwise joint intervention effects with their additive expectations. The joint effects deviate noticeably from the additive baseline, with both positive and negative interactions observed. Specifically, 53.3\% of factor pairs exhibit positive interactions, while 46.7\% exhibit negative interactions, with a mean absolute interaction magnitude of 0.084. The interaction map reveals heterogeneous cross-factor dependencies, including reinforcing and antagonistic effects, suggesting non-additive factor interactions under distribution shifts and providing complementary evidence for factor-level shift analysis.

\subsection{Factor-wise Shift Heterogeneity}

\begin{figure}[t]
    \centering
    \includegraphics[width=\linewidth]{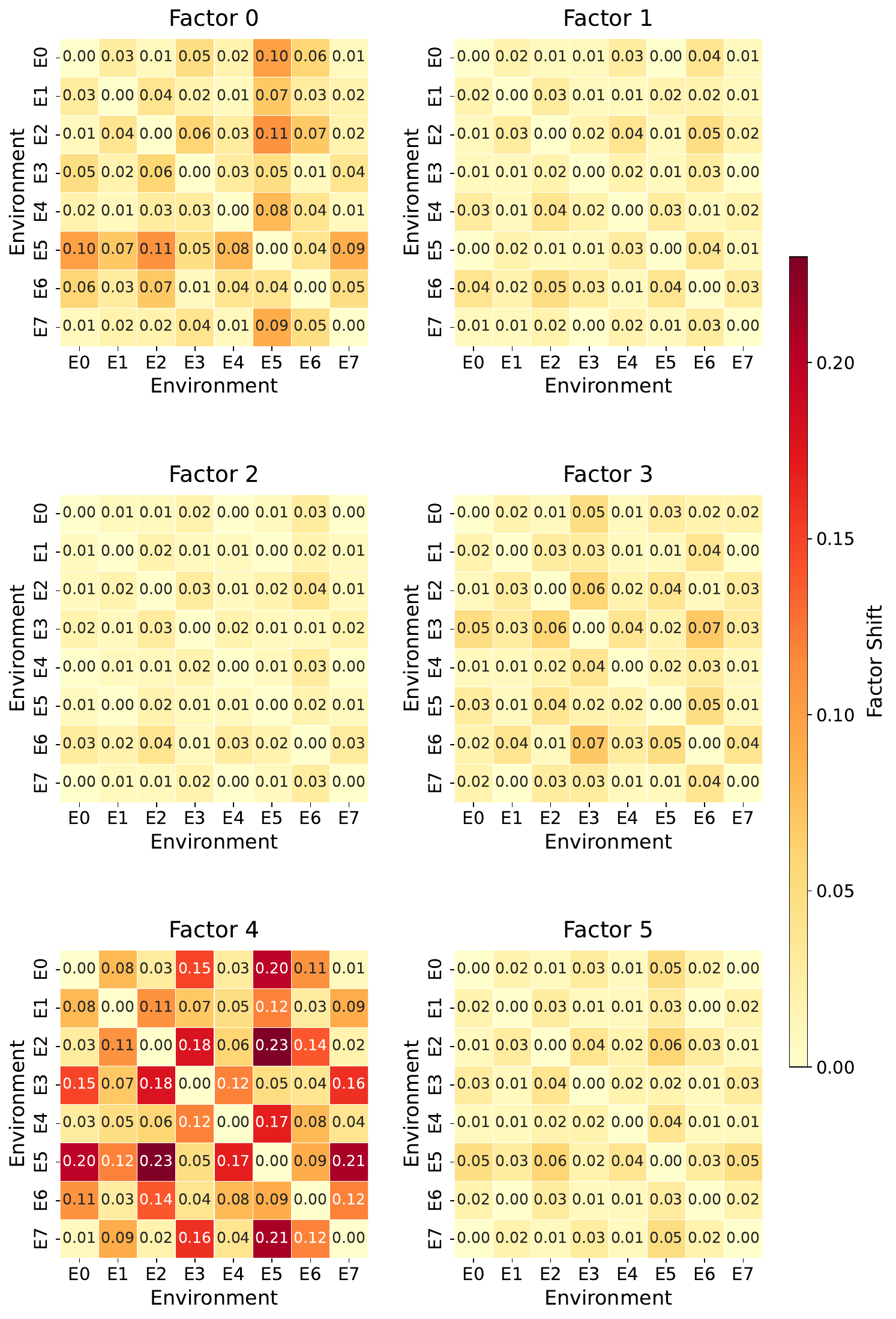}
    \caption{Cross-environment shift magnitude across the learned factors.}
    \label{fig:factor_shift_map}
\end{figure}

Figure~\ref{fig:factor_shift_map} shows substantial heterogeneity in cross-environment distribution shifts across the learned factors. $F_4$ exhibits the largest shift magnitude under speaker-based environment changes, indicating that its distribution is particularly sensitive to speaker-specific variations. Although the factors are not explicitly assigned semantic meanings, this pattern suggests that $F_4$ may encode speaker-dependent behavioral or expressive characteristics. $F_0$ also shows a relatively pronounced shift, whereas ${F_2,F_5}$ exhibit comparatively smaller changes across environments. These results reveal heterogeneous sensitivities of individual factors to speaker-induced shifts. Since shift magnitude alone cannot determine their predictive relevance or stability, we jointly consider shift magnitude, intervention effects, and cross-environment stability to guide selective factor treatment.

\section{Conclusion}
In this work, we challenge the conventional view that distribution shifts in multimodal learning should be universally suppressed. We introduce an intervention-aware shift framework that factorizes multimodal representations and characterizes each factor by its distributional shift, intervention effect, and cross-environment stability. Based on these properties, the framework selectively suppresses irrelevant shifts while preserving or adapting to task-relevant ones. Extensive experiments across multiple multimodal benchmarks demonstrate that FaShift improves generalization to unseen environments while retaining sensitivity to meaningful changes. These results advocate moving beyond invariance toward factor-wise selective invariance, where each shift is treated according to its predictive relevance and cross-environment stability.

\subsection*{Limitations}

Despite the effectiveness of FaShift, several limitations remain. First, the intervention-based analysis relies on latent factor replacement to estimate predictive sensitivity. Although this provides a practical way to characterize how factor perturbations affect model predictions, it does not establish formal causal effects or guarantee causal identifiability. Future work could incorporate explicit causal structures or stronger intervention mechanisms to further validate the causal interpretation of factor-level shifts. Second, the learned latent factors are not explicitly aligned with predefined semantic concepts, and their identifiability may depend on the factorization objective and model initialization. While our empirical results demonstrate their utility for shift analysis, a more principled characterization of factor semantics and identifiability would strengthen the framework. Third, some experiments rely on pseudo-environments constructed from learned factor representations to simulate distribution shifts. Although we additionally evaluate FaShift under naturally structured temporal shifts, broader evaluation across real-world environment changes would provide stronger evidence of its generality. Finally, the current treatment policy relies on empirically selected thresholds to distinguish predictive relevance and stability, and the joint factor interactions are currently used for analysis rather than directly incorporated into the treatment policy. Developing adaptive treatment criteria and explicitly modeling higher-order factor interactions are promising directions for future work.

\bibliography{acl_latex}

\appendix

\section{More Implementation Details}
\label{app:more_impl}
Our framework follows a \textbf{learn--diagnose--refine paradigm}. First, we learn a factorized multimodal representation by decomposing the joint representation into non-redundant latent factors and optimizing them for the target task. After training, the encoder and prediction head are frozen. We then perform \textbf{training-free shift diagnosis}: for each factor, we measure its distributional shift across environments using a distribution discrepancy (e.g., MMD), intervene on the shifted factor while keeping other factors fixed, and quantify its intervention-based predictive relevance by the resulting prediction change. The consistency of intervention effects across environment pairs is further used to assess cross-environment stability. Based on these profiles, shifted factors are selectively treated: \textbf{nuisance shifts are suppressed, stable consequential shift shifts are preserved, and unstable consequential shift shifts are adaptively refined}. This design avoids universal invariance and instead retains or adjusts environmental variations according to their relevance to the task. The main hyperparameter settings are summarized in Table \ref{tab:hyper_settings}.

In summary, Stage I requires training, whereas Stage II performs factor-wise shift diagnosis without additional parameter training. Stage III applies the resulting treatment rules without introducing an additional adaptation network. For fair comparison, all methods use the same data splits and evaluation protocols. Unless otherwise stated, each experiment is repeated with 10 random seeds, and we report the average performance.

\paragraph{Environment Construction.}
We instantiate environments according to the natural data-generating structure of each dataset. For MOSI, MOSEI and CH-SIMS, each speaker defines an environment, capturing speaker-specific linguistic, acoustic, and visual characteristics. \emph{These environment definitions provide observed distributional contrasts for subsequent factor-level shift diagnosis without using prediction labels to construct environments.} Formally, for speaker $u$, we define
\begin{equation}
E_u = \{x_{u,v}\}_{v=1}^{N_u},
\end{equation}
where $x_{u,v}$ denotes the $v$-th video sample of speaker $u$, and $N_u$ is the total number of video samples associated with that speaker. Environment pairs are constructed from distinct speakers. The speaker-level target labels are used only for prediction and are not involved in environment construction.

For MUD3, since exact timestamps are unavailable but videos from each user are chronologically ordered, we group every 10 consecutive videos from the same user into an environment. Formally, we define
\begin{equation}
E_{u,t} = \{x_{u,v} \mid 10(t-1) < v \leq \min(10t, N_u)\},
\end{equation}
where $x_{u,v}$ denotes the $v$-th video of user $u$ in chronological order, $N_u$ is the total number of videos associated with user $u$, and $t$ indexes the temporal environments. This construction approximates temporal variations in multimodal behavioral characteristics using the chronological order of video samples.


\paragraph{Design Principles for Factorization.}
Stage I does not aim to directly identify predictive factors. Instead, it learns a set of predictive, non-redundant, and intervenable latent factors. Stage II then leverages environmental variations and factor-level interventions to characterize their stability and shift properties. The factorization is guided by four principles:

\textbf{1. Task relevance.} The factorized representation should preserve the task-relevant information in the original multimodal representation, such that \(g(Z)\) maintains comparable predictive performance to \(g(H)\).

\textbf{2. Low redundancy.} Different factors should capture sufficiently distinct information. We therefore encourage low statistical dependence between factor pairs using the HSIC-based regularization.

\textbf{3. Intervenability.} Each factor should constitute a distinct representation unit that can be individually replaced or modified, enabling factor-level interventions for subsequent shift analysis.

\textbf{4. No predefined semantics.} We do not impose semantic labels or modality-specific meanings on individual factors. Each factor is learned from the downstream task and may capture an arbitrary predictive pattern emerging from the multimodal input.

\begin{table}[t]
\centering
\small
\setlength{\tabcolsep}{8pt}
\begin{tabular}{ccccccc}
\toprule
\textbf{Dataset} & \textbf{K} & \textbf{M} & $\tau_{s}$& $\tau_{c}$ & $\tau_{v}$ & $\lambda_{desc}$\\
\midrule
MOSI & 6 & 8 & 0.02 &0.2 &0.6 &500\\
MOSEI & 6 & 8  & 0.02&0.2 &0.6 &500\\
CH-SIMS& 8 & 4 &0.05 & 0.3 &0.6 &400\\
MUD3& 8 & 6 & 0.05&0.3 &0.6 &200\\
\bottomrule
\end{tabular}
\caption{Hyperparameter settings of FaShift across different multimodal datasets.}
\label{tab:hyper_settings}
\end{table}

\paragraph{Reference Factor Construction.}
The reference factor $Z_k^{a,*}$ is obtained by averaging the
corresponding $Z_k$ representations of the selected counterparts:
\begin{equation}
Z_k^{a,*}=
\frac{1}{M}
\sum_{j\in\mathcal{N}_k^a(b)} Z_k^j,
\end{equation}
where $\mathcal{N}_k^a(b)$ denotes the set of top-$M$ matched samples
from $E_a$. This averaged factor is used to replace $Z_k^b$ while
keeping all non-intervened factors $Z_{-k}^b$ unchanged.

\begin{table}[t]
\centering
\small
\setlength{\tabcolsep}{3.5pt}
\begin{tabular}{lcccc}
\toprule

& \multicolumn{2}{c}{\textbf{ID}}
& \multicolumn{2}{c}{\textbf{OOD}} \\
\cmidrule(lr){2-3}
\cmidrule(lr){4-5}
& \textbf{Acc.} & \textbf{F1}
& \textbf{Acc.} & \textbf{F1} \\
\midrule
\cite{wang2025beyond}
& 78.3 & 77.3 & 70.9 & 70.7 \\
\textbf{FaShift(ours)}
& 82.4 & 82.7 & 74.1 & 74.4 \\
\bottomrule
\end{tabular}
\caption{Results of MUD3 datasets.}
\label{tab:appendix_table}
\end{table}

\paragraph{Factorized Representation.}
We use a pretrained multimodal encoder (Qwen3-Embedding-4B\footnote{https://huggingface.co/Qwen/Qwen3-Embedding-4B}) to obtain a joint representation $H\in\mathbb{R}^{d}$, which is further decomposed into $K$ compact
factors, with each factor having dimension $d_f$. In our experiments,
we set $K=6$ and $d_f=32$. The factorization module is trained jointly
with the prediction head, while the resulting factorized representation is fixed during the subsequent factor-wise shift analysis.

\paragraph{Distribution Shift Estimation.}
For each environment $E_a$, we collect the empirical samples of the
$k$-th factor as:
\begin{equation}
\mathcal{Z}_k^a
=
\{z_{k,i}:E_i=E_a\}.
\end{equation}
We then quantify the cross-environment shift using MMD (Maximum Mean Discrepancy):
\begin{equation}
D_k^{a,b}
=
\operatorname{MMD}
\left(
\mathcal{Z}_k^a,\mathcal{Z}_k^b
\right).
\end{equation}
We quantify the distributional shift of each factor using the maximum
mean discrepancy (MMD) between its empirical distributions across
environments. For each factor $z_k$, we compare its representations
from a pair of environments and aggregate the pairwise MMD scores
across all environment pairs.

\paragraph{Counterpart Matching.}
For each target sample from environment $E_b$ and factor $z_k$,
we retrieve the top-$M$ nearest samples from the reference environment
$E_a$ based on the similarity of their non-intervened representations
$z_{-k}$. Specifically, we use cosine similarity to measure the
similarity between the target representation and each candidate in
$E_a$, and select the $M$ candidates with the highest similarity.

\paragraph{Counterfactual Intervention.}
All counterfactual interventions are performed without updating model
parameters. For each shifted factor, we replace only the target factor
with its reference-environment counterpart while keeping all remaining
factors unchanged. The factual and counterfactual samples are then
passed through the same prediction head to measure the intervention-
induced prediction change.

\paragraph{Intervention Effect.}
We quantify the intervention effect $\Delta_k$ as the $\ell_1$ distance between the factual and counterfactual prediction probability vectors. This measure captures the magnitude of prediction changes induced by intervening on factor $F_k$, providing a finer-grained assessment of its predictive influence than binary label flipping.

\begin{table}[t] \centering \small \setlength{\tabcolsep}{2pt} \begin{tabular}{llcccc} \toprule \textbf{Dataset} & \textbf{Environment} & \multicolumn{2}{c}{\textbf{ID}} & \multicolumn{2}{c}{\textbf{OOD}} \\ & & Acc-2 $\uparrow$ & F1 $\uparrow$ & Acc-2 $\uparrow$ & F1 $\uparrow$ \\ \midrule \multirow{2}{*}{MOSEI} & Observed & 88.2 & 88.3 & 78.5 & 78.4 \\ & Pseudo & 82.5 & 82.6 & 75.7 & 74.5 \\ \midrule \multirow{2}{*}{CH-SIMS} & Observed &83.3 & 83.3 & 75.4 & 75.2 \\ & Pseudo & 79.1 & 79.6 & 70.6 & 70.3 \\ \bottomrule \end{tabular} 
\caption{Comparison between observed and pseudo-environments for factor-wise shift diagnosis. For pseudo-environments, observed environmental attributes are excluded, and environments are constructed solely from Stage-I factor activation profiles.} \label{tab:pseudo_env} 
\end{table}

\section{More Results}
\label{app:more_results}

\paragraph{Results under Temporal Shifts.}

Table~\ref{tab:appendix_table} presents the performance of FaShift on MUD3 under different environment settings. For the MUD3 dataset, we construct environments according to recording time. Specifically, samples are chronologically partitioned into distinct temporal groups, with each group treated as an environment. This construction captures temporal variations in multimodal characteristics and induces realistic distribution shifts across environments, providing a natural setting for evaluating model robustness to temporal changes.

FaShift consistently outperforms the compared method in all evaluation metrics, demonstrating improved predictive performance in both in-distribution and distribution-shifted settings. Notably, FaShift maintains a clear advantage under OOD evaluation, indicating that its factor-wise shift treatment remains effective when multimodal characteristics vary across environments. These results further support the generalizability of FaShift to temporally structured distribution shifts beyond the benchmark datasets considered in the main experiments.

\begin{figure}[t]
    \centering
    \includegraphics[width=\linewidth]{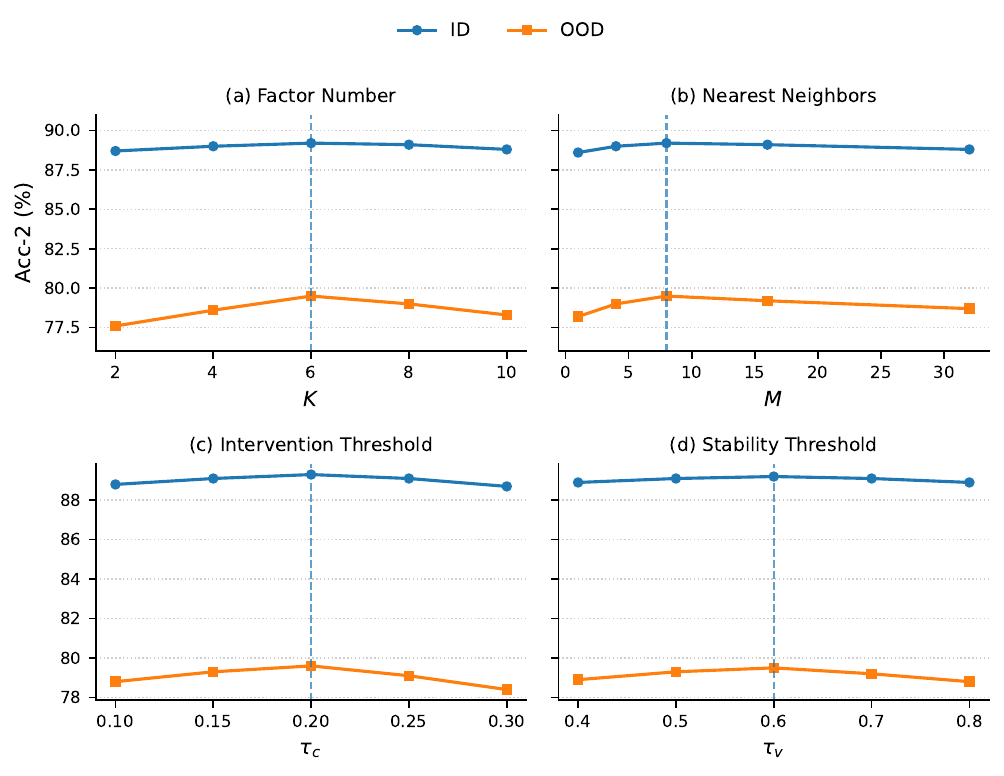}
    \caption{Hyperparameter sensitivity of FaShift under ID and OOD settings. We vary the number of nearest neighbors $M$, intervention-effect threshold $\tau_c$, and stability threshold $\tau_v$, while keeping other hyperparameters fixed.}
    \label{fig:hyperparameter}
\end{figure}

\paragraph{Pseudo-Environment Analysis.}
Pseudo-environments provide an alternative operational grouping when reliable environmental attributes are unavailable. Table~\ref{tab:pseudo_env} evaluates FaShift using representation-derived pseudo-environments without relying on observed environmental attributes. Although pseudo-environments result in a moderate performance drop compared with naturally defined environments, FaShift maintains competitive OOD performance on both MOSEI and CH-SIMS v2. This indicates that factor-wise shift diagnosis and adaptive treatment remain effective when explicit environmental attributes are unavailable. The performance gap further suggests that naturally defined environments provide more informative heterogeneity, while pseudo-environments offer a viable alternative for datasets lacking reliable environmental attributes.

\paragraph{Hyperparameter Sensitivity.}

\begin{table}[t]
\centering
\small
\setlength{\tabcolsep}{8pt}
\begin{tabular}{ccc}
\toprule
\textbf{$K$} & \textbf{Avg. Factor Corr} & \textbf{OOD Acc-2 (\%)} \\
\midrule
2 & 0.65 & 74.6 \\
4 & 0.35 & 76.8 \\
\textbf{6} & \textbf{0.15} & \textbf{79.5} \\
8 & 0.38 & 75.1 \\
\bottomrule
\end{tabular}
\caption{Sensitivity of FaShift to the number of latent factors $K$ on MOSEI.}
\label{tab:factor_number_sensitivity}
\end{table}

Figure~\ref{fig:hyperparameter} evaluates the sensitivity of FaShift to the key hyperparameters $M$, $\tau_c$, and $\tau_v$, while Table~\ref{tab:factor_number_sensitivity} investigates the effect of the number of latent factors $K$. The performance remains relatively stable across a broad range of values under both ID and OOD settings, indicating that FaShift is not highly sensitive to the choice of hyperparameters. In particular, the selected values of $M=8$, $\tau_c=0.2$, and $\tau_v=0.6$ achieve consistently strong performance without requiring fine-grained tuning. These results suggest that the proposed factor-wise shift diagnosis and adaptive treatment are robust to moderate variations in the intervention and stability criteria.

\end{document}